\pdfoutput=1

\documentclass[11pt]{article}

\usepackage[]{ACL2023}

\usepackage{times}
\usepackage{latexsym}
\usepackage{graphicx}
\usepackage{multirow}
\usepackage{booktabs}
\usepackage{hyperref}
\usepackage{enumitem}

\usepackage[T1]{fontenc}

\usepackage[utf8]{inputenc}

\usepackage{microtype}

\usepackage{inconsolata}

\usepackage[para]{footmisc}

\newcommand{\taco}{\textsc{TacoBot}}

\title{Roll Up Your Sleeves: Working with a Collaborative and Engaging \\Task-Oriented Dialogue System}

\author{
    Lingbo Mo, Shijie Chen\Thanks{Team co-leads in the challenge with equal contribution. } , Ziru Chen\footnotemark[\value{footnote}] , Xiang Deng\Thanks{Other authors in alphabetical order.} , Ashley Lewis\footnotemark[\value{footnote}] , Sunit Singh\footnotemark[\value{footnote}] \\ 
    \textbf{Samuel Stevens\footnotemark[\value{footnote}] , Chang-You Tai\footnotemark[\value{footnote}] , Zhen Wang\footnotemark[\value{footnote}] , Xiang Yue\footnotemark[\value{footnote}] , Tianshu Zhang\footnotemark[\value{footnote}] , Yu Su\Thanks{Faculty advisors.} , Huan Sun\footnotemark[\value{footnote}]}\\	
    The Ohio State University\\
	\texttt{\{mo.169, chen.10216, chen.8336, deng.595, lewis.2799, singh.1790,} \\
	\texttt{stevens.994, tai.97, wang.9215, yue.149, zhang.11535, su.809, sun.397\}@osu.edu}
}

\begin{document}
\maketitle

\begin{abstract}

We introduce \taco{}, a user-centered task-oriented digital assistant designed to guide users through complex real-world tasks with multiple steps. Covering a wide range of cooking and how-to tasks, we aim to deliver a collaborative and engaging dialogue experience. Equipped with language understanding, dialogue management, and response generation components supported by a robust search engine, \taco{} ensures efficient task assistance. To enhance the dialogue experience, we explore a series of data augmentation strategies using LLMs to train advanced neural models continuously. \taco{} builds upon our successful participation in the inaugural Alexa Prize TaskBot Challenge, where our team secured third place among ten competing teams. We offer \taco{} as an open-source framework that serves as a practical example for deploying task-oriented dialogue systems.\footnote{Code and datasets are available at \href{https://github.com/OSU-NLP-Group/TacoBot}{OSU-NLP/TacoBot}.}

\end{abstract}
\section{Introduction}

Task-Oriented Dialogue (TOD) systems have shown promise in achieving user goals through conversational interactions~\citep{SMDataflow2020, su2022multi, mo2022towards}. However, existing TOD systems focus on users providing information while the system performs tasks. In contrast, our task bot assists users in executing tasks themselves by providing accurate information and guidance.

\begin{figure}[t]
  \centering
  \includegraphics[width=1\columnwidth]{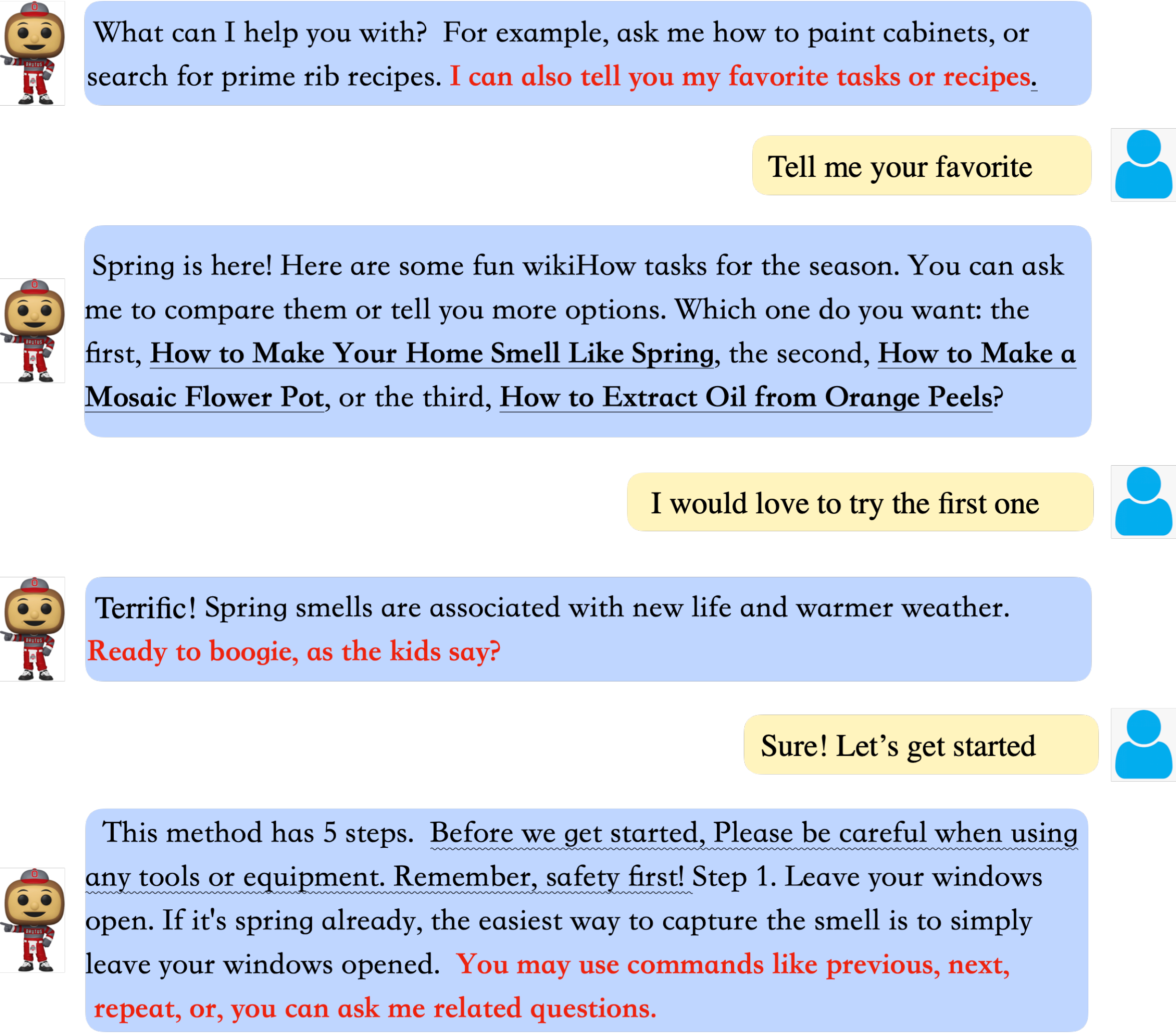}
  \caption{An example dialogue showing first few turns.}
  \vspace{-10pt}
  \label{dialogue_example}
\end{figure}

However, we face several challenges, including the following: (1) Existing TOD systems prioritize functional goals at the expense of user experience. (2) Inadequate in-domain training data, as modern neural models require large amounts of data, and acquiring annotations through crowdsourcing is costly. In this paper, we present \taco{}, a task-oriented dialogue system designed to assist users in completing multi-step cooking and how-to tasks. Built upon our previous bot~\citep{chen2022bootstrapping} deployed in the Alexa Prize TaskBot Challenge~\citep{gottardi2022alexa}, \taco{} aims to deliver a collaborative and engaging user experience. Figure~\ref{dialogue_example} showcases a partial example dialogue.

Our contributions include: (1) Developing a modularized TOD framework with accurate language understanding, flexible dialogue management, and engaging response generation. (2) Exploring data augmentation strategies, such as leveraging GPT-3 to synthesize large-scale training data. (3) Introducing clarifying questions about nutrition for cooking tasks to personalize search and better cater to user needs. (4) Incorporating chit-chat functionality, allowing users to discuss open topics of interest beyond the task at hand.

\section{System Design}

\subsection{System Overview}

\taco{} follows a canonical pipeline approach for TOD systems. The system consists of three main modules: Natural Language Understanding (NLU), Dialogue Management (DM), and Response Generation (RG). NLU module preprocesses the user's utterance to determine their intent. DM module, designed with a hierarchical finite state machine, controls the dialogue flow, handles exceptions, and guides the conversation towards task completion. RG module generates responses using relevant knowledge and additional modalities to enhance user engagement. Each module is supported by a well-organized knowledge backend and search engine, capable of connecting with various sources to provide optimal user assistance.

\subsection{Natural Language Understanding}
\label{nlu}

\begin{table*}[ht]
\centering
\small
\begin{tabular}{p{1.5cm}|p{13cm}}
\hline
\textbf{Category} & \textbf{Description} \\
\hline
Sentiment & The user can confirm or reject the bot's response on each turn, leading to three labels: \texttt{Affirm}, \texttt{Negate}, and \texttt{Neutral}, indicating the user utterance's polarity. \\
\hline
Commands & The user can drive the conversation using these commands: \textbf{Task Request}, \textbf{Navigation} (to view candidate tasks or walk through the steps), \textbf{Detail Request}, \textbf{PAK Request}, \textbf{Task Complete}, and \textbf{Stop} to terminate the conversation at any time. \\
\hline
Utilities & We use a \textbf{Question} intent to capture user questions and a \textbf{Chat} intent for casual talk. \\
\hline
Exception & To avoid unintentional changes in dialogue states, we have one additional intent for out-of-domain inputs, such as incomplete utterances and greetings. \\
\hline
\end{tabular}
\caption{Categories of detailed intents to support diverse user initiatives.}
\label{intent_cat}
\end{table*}

Our bot employs a robust NLU pipeline which fuses the strengths of pre-trained language models with rule-based approaches. The key component is \textit{Intent Recognition}, where we organize multiple intents into four categories to accommodate  a wide array of user initiatives, as detailed in Table~\ref{intent_cat}. Real-world user initiatives often encompass several intents within one single utterance. Accordingly, we address intent recognition as a multi-label classification problem and filter model predictions according to the dialogue state.

To develop a high-quality multi-label classification model despite limited data, we employ data augmentation and domain adaptation techniques. We leverage existing datasets~\citep{dstc8} for common intents like \textbf{Sentiment} and \textbf{Question}, while utilizing the in-context learning capability of GPT-3 for other intents. By synthesizing initial utterances with intent descriptions and few-shot examples, we create a foundation for training data. To expand the dataset, we transform synthetic utterances into templates, substituting slot values with placeholders and filling them with sampled values to generate actual training utterances. Additionally, we incorporate linguistic rules, neural paraphrase models, and user noise, such as filler words, to enhance data diversity and improve the robustness of our intent recognition module.


\subsection{Dialogue Management}

\begin{figure}[t]
  \centering
  \includegraphics[width=0.9\columnwidth]{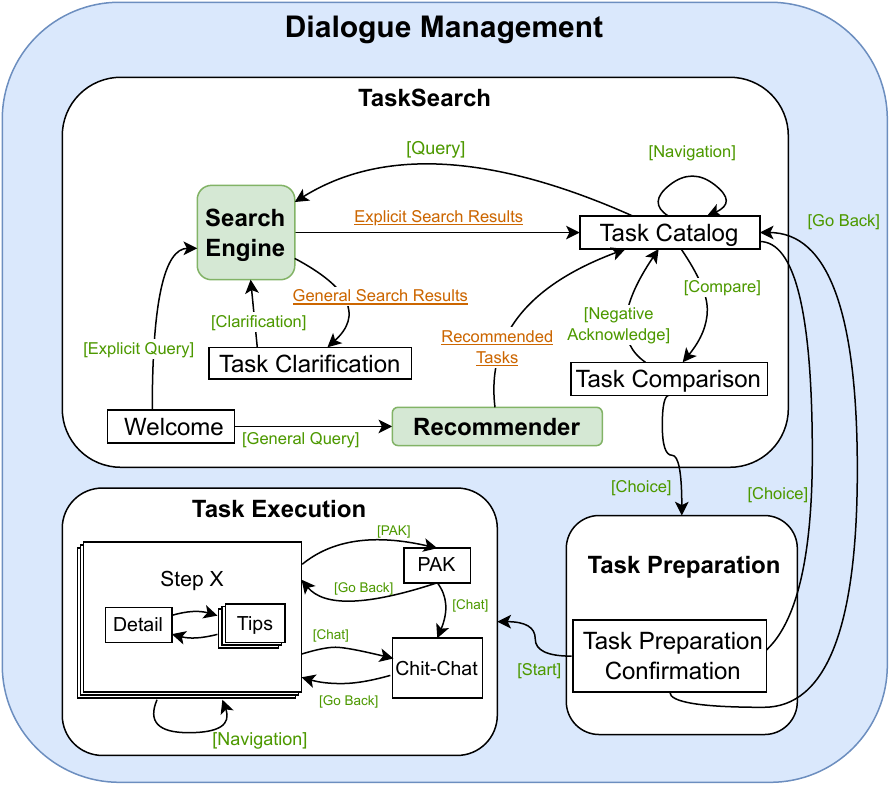}
  \caption{Dialogue Management Diagram. White boxes represent dialogue states and green boxes represent supporting modules. Bidirectional edges represent reflexive transitions. Green texts represent user intent and orange texts denote search engine output.}
  \label{dm}
\end{figure}

We design a hierarchical finite state machine for the DM component, consisting of three phases: Task Search, Task Preparation, and Task Execution. Each phase comprises multiple fine-grained dialogue states, as depicted in Figure~\ref{dm}.

In the \textbf{Task Search phase}, users can search for how-to tasks or recipes directly by issuing a query or ask for task recommendations. \taco{} retrieves search results from the backend search engine (Section~\ref{search}) and presents candidate tasks for users to compare and select. Once users choose an option, they enter the \textbf{Task Preparation phase}. In this phase, users review detailed information about the selected task and decide whether to proceed or search for another task. If users change their mind, they can go back to Task Search and find an alternative task. If they commit to the chosen task, they proceed to the \textbf{Task Execution phase}. During this last phase, users follow the step-by-step instructions provided by \taco{} to complete the task. The utility module, such as the QA module, assists users throughout this phase. Each step of the task has its own state, and for how-to tasks, we break down lengthy steps into shorter instructions, details, and tips for better user comprehension.

DM performs state transitions and selects response generators (Section~\ref{rg}) based on user input. The hierarchical design of dialogue states allows for extensible and flexible transitions at different levels. A dialogue state history stack is maintained to facilitate easy navigation to previous states. User intents that do not trigger valid transitions provide contextualized help information to guide users through the dialogue. These design choices ensure stable yet flexible dialogue experiences for users.


\subsection{Search Engine}
\label{search}

\taco{} can support diverse tasks backed by large-scale corpus. For the cooking domain, we build a recipe corpus which contains 1.02M recipes based on Recipe1M+ dataset~\citep{marin2021recipe1m+}. Meanwhile, we build a wikiHow corpus that includes 93.1K how-to tasks collected from wikiHow website\footnote{\url{https://www.wikihow.com}}. On top of that, we construct a search engine for both domains based on Elastic search.  

\subsubsection{Ranking Strategy}

To improve the relevance of search results and mitigate the issue of lexical similarity in Elastic search, we employ a query expansion technique that expands user queries by incorporating related words from task names, such as lemmatized verbs, nouns, and decomposed compound nouns. Additionally, we enhance search performance by implementing a neural re-ranking model based on BERT. This model assigns a score to each task by considering the task request and retrieved task titles as input. Training the re-ranker involves employing a weakly-supervised list-wise ranking loss and utilizing synthesized task queries via GPT-3 query simulation. We also propose the collection of weak supervision signals from Google's search engine to avoid the need for human annotation.

\subsubsection{Personalized Search}

\begin{figure}[h]
  \centering
  \includegraphics[width=0.9\columnwidth]{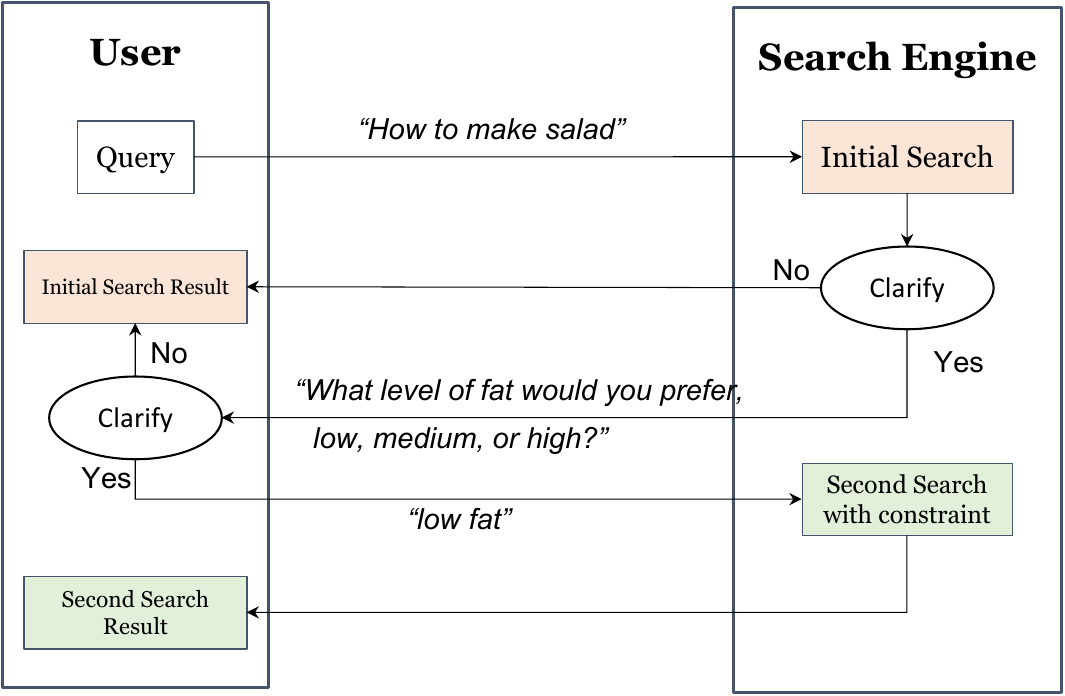}
  \caption{The flow chart for asking clarifying questions.}
  \label{clarifying}
\end{figure}

 In addition to implementing ranking strategies for accurate search results, our goal is to infuse personalization into the search engine, ensuring a more finely-tuned match with users' needs. To achieve this, we propose a method of asking clarifying questions during recipe searches, collaborating closely with users to understand their preferences regarding nutrition. The logic flow of the process is depicted in Figure~\ref{clarifying}. Specifically, when a user provides a cooking task of interest, we proactively engage in clarifying discussions with them about the desired level of nutrition in terms of \textit{sugar}, \textit{fat}, \textit{saturates}, and \textit{salt}, using the traffic lights definition established by the Food Standards Agency (FSA).

\subsection{Response Generation}
\label{rg}

Our response generation module blends both infilling-based methods and neural models. We leverage handcrafted conditional rules to organize curated templates and their composition strategy according to the high-level states in our hierarchical finite-state machine. Simultaneously, we build a QA system to respond to diverse user queries.

\subsubsection{Question Type Classifier}

Our QA system encompasses various question types, including in-context machine reading comprehension (MRC) for context-dependent questions, out-of-context (OOC) QA for open domain questions, frequently-asked questions (FAQ) retrieval for how-to tasks, and rule-based Ingredient and Substitute QA for cooking tasks.

Then, we develop a question type classifier that categorizes user questions into five types (MRC, OOC, FAQ, Ingredient, Substitute) for cooking tasks, and three types (MRC, OOC, FAQ) for how-to tasks. To improve classification accuracy, we concatenate the instruction of the current step (if available) as context with the input question. This combined sequence is then fed into a Roberta-base classifier. Our training set consists of 5,000 questions for each question type, allowing for effective differentiation between different types of questions.


\subsubsection{Context-Dependent QA}

We begin by annotating an in-context QA dataset comprising 5,183 QA pairs, out of which 752 are unanswerable questions. To ensure reliable responses, we employ Roberta-base to build an extractive QA model in two stages. Initially, we pre-train our model on SQuAD 2.0, followed by fine-tuning on our annotated QA dataset. Recognizing that users may inquire about previously shown steps, we enhance the context by concatenating the current step with the preceding $n$ steps ($n=2$) during both training and inference processes to prevent information gaps and hallucination.

\subsubsection{Context-Independent QA}

\taco{} supports both in-context and context-independent questions. For \textbf{out-of-context QA}, we utilize \textsc{FLAN-T5-XXL}~\citep{chung2022scaling}, an instruction-finetuned language model with 11B parameters. Under the zero-shot prompting setup, our bot is equipped to handle open-domain QA and demonstrate commonsense reasoning. 

Additionally, \textbf{FAQ} module leverages common questions from wikiHow's Community Q\&A section, providing answers sourced from real user questions and expert responses. We use a retrieval module based on cosine similarity with question embeddings generated by a sentence-BERT encoder. For \textbf{ingredient-related queries}, we employ a high-recall string matching mechanism against the recipe's ingredient list. If users lack a specific ingredient, we suggest alternatives, leveraging a dataset covering 200 commonly used ingredients.


\subsection{User Engagement}

We develop several strategies to pursue an engaging dialogue experience in the following sections.

\subsubsection{Chit-Chat}

In real-world conversations, users often desire casual talk alongside the task. To enhance the user experience, \taco{} offers chit-chat functionality, enabling flexible and diverse conversations. Inspired by Chirpy Cardinal~\citep{chi2022neural}, we integrate a chit-chat module into our TOD system. A template-based strategy is employed to identify user intent when entering and exiting chit-chat. The chit-chat process consists of three components.

Firstly, \textbf{Entity Tracker} monitors entities throughout the conversation, aligning responses with user intentions and focusing on the current topic. Recognized entities allow \taco{} to access web sources (Wikipedia and Google) and provide intriguing information. Secondly, \textbf{Chit-Chat Response Generator} incorporates various response generators: Neural Chat, Categories, Food, Aliens, Wiki, and Transition. Neural Chat uses BlenderBot-3B to generate open-domain responses. Categories and Food generators elicit entity-related responses using templates. Transition facilitates smooth shifts between entities. Wiki enables users to discover engaging information in a conversational style. Aliens presents a five-part monologue series on extraterrestrial existence. Lastly, \textbf{Intent Identification Model} determines if the user wants to continue or shift topics. \taco{} proactively prompts users to return to the task after some chit-chat. Achieving natural transitions between chit-chat and task-oriented dialogue requires ongoing efforts.


\subsubsection{People Also Ask}

Furthermore, \taco{} aims to enhance the dialogue experience by delivering captivating content. We leverage Google's ``People Also Ask'' (PAK) feature, which provides a list of related questions and summarized answers from web pages. This feature reveals popular topics of interest. To collect PAK data, we extract 30k common keywords from task titles in our recipe and wikiHow corpus, resulting in a total of 494k PAK QA pairs.

During task execution, PAK is presented as additional information. To avoid disrupting user focus, we limit the display frequency, currently showing it every 3 steps. Instead of directly displaying the PAK QA pair, we offer an interactive experience by presenting the question first, allowing users to decide if they want to view the corresponding answer. We also provide the option for users to engage in chit-chat if they choose to view PAK.


\section{Conclusion}
\label{conclusion}

In this paper, we introduce \taco{}, a modular task-oriented dialogue system that assists users in accomplishing intricate daily tasks. We propose a comprehensive set of modules and approaches to create a collaborative and engaging task bot. To ensure a strong foundation, we employ several data augmentation techniques leveraging LLMs. Furthermore, we open-source the framework and datasets, providing a valuable resource and inspiring future efforts to enhance user-bot collaboration.

\section*{Ethics Statement}

We present a task bot that is able to converse with users to complete real-world tasks. No personal or identifying information is included throughout conversations. In addition, our bot includes a safety check to ensure safe conversations. We reject inappropriate task requests and prevent showing dangerous tasks, where users and their properties may get hurt. To this end, we perform rule-based matching against a keyword blacklist to filter out inappropriate tasks. Meanwhile, for response generation, we don't directly use LLMs (such as ChatGPT) to generate answers for users' questions, which will have the risk of leaking user data to third-party APIs. Instead, we utilize LLMs to do data augmentation and domain adaptation, and train models locally for the sake of privacy protection.

\section*{Author Contributions}

In this work, each author makes significant contributions that collectively enhance the final outcome. Lingbo Mo played a crucial role in constructing and organizing the codebase, and building an interactive interface for the demo. Shijie Chen and Ziru Chen co-led the team during the challenge, laying the groundwork for the bot's development. Xiang Deng and Tianshu Zhang were responsible for the NLU pipeline. Xiang Yue mainly developed the QA module. Lingbo Mo and Zhen Wang worked together to build the backend knowledge base and the search engine. Samuel Stevens provided engineering support for constructing an automated test suite. Ashley Lewis focused on enhancing user engagement. Chang-You Tai contributed to chit-chat and PAK features, while Sunit Singh assisted in designing the demo interface. Huan Sun and Yu Su are faculty advisors and offered valuable guidance and feedback.

\section*{Acknowledgements}

We thank colleagues in the OSU NLP group and Amazon for their valuable feedback. Part of the work was done during the first Alexa Prize TaskBot challenge and supported by Amazon.com, Inc. The work was also partly supported by NSF CAREER \#1942980.

\bibliography{anthology,custom}
\bibliographystyle{acl_natbib}




\end{document}